\documentclass[final,3p]{elsarticle}

\usepackage{lineno}

\usepackage{time}
\usepackage{epsfig}
\usepackage{graphicx}
\usepackage{amsmath}
\usepackage{amssymb}
\usepackage{booktabs}
\usepackage{color}
\usepackage{overpic}
\usepackage{wrapfig}
\usepackage{epstopdf}
\usepackage{pgf,tikz}
\usepackage{pgfplots}
\usepackage{tabulary}
\usepackage{multirow}
\usepackage{tabu}
\usepackage{sidecap}

\usepackage[colorlinks=true,breaklinks=true,pdftex]{hyperref}
\modulolinenumbers[1]

\journal{arXiv}

\biboptions{authoryear}

\newcommand{\x}{\mathbf{x}}

\newcommand{\zero}{\mathbf{0}}

\newcommand{\X}{\mathbf{X}}

\newcommand{\F}{\mathbf{F}}
\newcommand{\fb}{\mathbf{f}}

\newcommand{\svec}{\mathbf{s}}

\newcommand{\vb}{\mathbf{v}}
\newcommand{\nb}{\mathbf{n}}

\newcommand{\Sp}{\mathbb{S}}

\newcommand{\R}{\mathbb{R}}
\newcommand{\Rot}{\mathbf{R}}
\newcommand{\rb}{\mathbf{r}}

\newcommand{\Pb}{\mathbf{P}}

\usepackage{cleveref}

\Crefname{assumption}{\textbf{H}\hspace{-3pt}}{\textbf{H}\hspace{-3pt}}
\crefname{algorithm}{\text{Alg.}}{\text{Alg.}}
\crefname{assumption}{\textbf{H}}{\textbf{H}}
\crefname{equation}{\text{Eq}}{\text{Eq}}
\crefname{definition}{\text{Dfn.}}{\text{Dfn.}}
\crefname{lemma}{\text{Lemma}}{\text{Lemma}}
\crefname{dfn}{\text{Dfn.}}{\text{Dfn.}}
\crefname{thm}{\text{Thm.}}{\text{Thm.}}
\crefname{tab}{\text{Tab.}}{\text{Tab.}}
\crefname{fig}{\text{Fig.}}{\text{Fig.}}
\crefname{table}{\text{Tab.}}{\text{Tab.}}
\crefname{figure}{\text{Fig.}}{\text{Fig.}}
\crefname{section}{\text{Sec.}}{\text{Sec.}}

\newenvironment{packed_itemize}
{\begin{itemize}
    \setlength{\itemsep}{1pt}
    \setlength{\parskip}{0pt}
    \setlength{\parsep}{0pt}
}{\end{itemize}}

\begin{document}

\newcolumntype{x}[1]{>{\centering\arraybackslash}p{#1pt}}
\newlength\savewidth\newcommand\shline{\noalign{\global\savewidth\arrayrulewidth
  \global\arrayrulewidth 1pt}\hline\noalign{\global\arrayrulewidth\savewidth}}
\newcommand\hshline{\noalign{\global\savewidth\arrayrulewidth
  \global\arrayrulewidth 0.6pt}\hline\noalign{\global\arrayrulewidth\savewidth}}
\newcommand{\tablestyle}[2]{\setlength{\tabcolsep}{#1}\renewcommand{\arraystretch}{#2}\centering\footnotesize}
\makeatletter\renewcommand\paragraph{\@startsection{paragraph}{4}{\z@}
  {.5em \@plus1ex \@minus.2ex}{-.5em}{\normalfont\normalsize\bfseries}}\makeatother
\newcommand{\mr}[1]{\multirow{2}{*}{#1}}

\newcommand{\todo}[1]{{\textcolor{red}{\textbf{#1}}}}
\newcommand{\orl}[1]{{\textcolor{red}{\textbf{Or}: #1}}}
\newcommand{\tolga}[1]{{\textcolor{purple}{\textbf{tolga}: #1}}}
\newcommand{\leo}[1]{{\textcolor{green}{\textbf{Leo}: #1}}}
\newcommand{\srinath}[1]{{\textcolor{magenta}{\textbf{Srinath}: #1}}}
\newcommand{\george}[1]{{\textcolor{cyan}{\textbf{George}: #1}}}
\newcommand{\ourspg}{PN}
\newcommand{\ourscg}{CC}

\begin{frontmatter}

\title{Continuous Geodesic Convolutions for Learning on 3D Shapes}


\author[first]{Zhangsihao Yang}
\ead{zhangsiy@cs.stanford.edu}

\author[second]{Or Litany}
\ead{or.litany@gmail.com}

\author[third]{Tolga Birdal}
\ead{tbirdal@stanford.edu}

\author[fourth]{Srinath Sridhar}
\ead{ssrinath@cs.stanford.edu}

\author[last]{Leonidas Guibas\corref{cor}}
\ead{guibas@cs.stanford.edu}

\cortext[cor]{Corresponding author}

\address[first, second, third, fourth, last]{Stanford University}

\begin{abstract}

The majority of descriptor-based methods for geometric processing of non-rigid shape rely on hand-crafted descriptors. Recently, learning-based techniques have been shown effective, achieving state-of-the-art results in a variety of tasks. Yet, even though these methods can in principle work directly on raw data, most methods still rely on hand-crafted descriptors at the input layer. In this work, we wish to challenge this practice and use a neural network to learn descriptors directly from the raw mesh. To this end, we introduce two modules into our neural architecture. The first is a local reference frame (LRF) used to explicitly make the features invariant to rigid transformations. The second is continuous convolution kernels that provide robustness to sampling. We show the efficacy of our proposed network in learning on raw meshes using two cornerstone tasks: shape matching, and human body parts segmentation. Our results show superior results over baseline methods that use hand-crafted descriptors. 
\end{abstract}


\end{frontmatter}


\section{Introduction}
Shape descriptors are key to many applications in 3D computer vision and graphics. Examples include shape matching, segmentation, retrieval, and registration, to name a few.
A good local descriptor should balance between two opposite forces. On the one hand, it needs to be discriminative enough to uniquely describe a surface local region. At the same time though, it needs to keep robustness to nuisance factors like noise or sampling. Depending on the task, other properties may be required. Common examples are invariance to rigid transformations~\citep{deng20193d,deng2018ppffold,deng2018ppfnet} or isometric deformations~\citep{masci2015geodesic, boscaini2016learning, monti2017geometric}. To this end, many descriptors have been manually crafted with built-in invariance. However, these rely on one's ability to analytically model structure in the data which can often be too difficult. Alternatively, machine learning approaches and neural networks in particular, can recover complex patterns from training samples. Further, end-to-end learning is task aware and thus can tailor the learned descriptors to the specified task. Neural networks have proven very powerful in learning descriptors from raw data in various domains including images, text, audio and point clouds. Recently, an exciting research branch termed geometric deep learning has emerged, offering various techniques to process shape represented as meshes. Interestingly though, the vast majority of methods still rely on hand-crafted descriptors at the input to the network as these seem to perform better than working on raw data. In this work, we wish to challenge this practice and learn directly from the raw mesh.  
Our proposed method is data-driven in nature, however, we integrate the powerful local reference frame (LRF) module commonly used in hand-crafted descriptors into our network.  We found through experimentation that structuring the learning through the LRF, is key to reach good performance.  A second contribution is the use of continuous convolution kernels. This concept was recently shown to be quite powerful in point cloud networks~\cite{wang2018deep}. Here, we show its usefulness in the context of deformable meshes. We show the efficacy of our proposed network in learning on raw meshes using two cornerstone tasks: shape matching, and human body parts segmentation. Our results show superior results over baseline methods that use hand-crafted descriptors. 
\paragraph{Contributions}
Our contributions can be summarized as follows.
\begin{packed_itemize}
    \item We introduce a local reference frame (LRF) and continuous convolution kernel modules in the context of deformable shapes.
    \item Using these, we are able to work directly on raw mesh features and outperform previous methods that take hand-crafted descriptors as input. 
    \item We achieve improved results on deformable shape matching, and human body part segmentation. 
\end{packed_itemize}

\section{Related Work}

\subsection{Shape descriptors}
\paragraph{Rigid Case} 
Rotation invariant 3D local descriptors are of great interest in the realm of 3D computer vision. Most of the works consider the scenario where the 3D point sets undergo a rigid transformation. The first handcrafted family of works tried to achieve repeatability under those transformations by building certain invariances such as isometry invariance~\citep{tombari2010unique,spin,rops,ppfh,fpfh,usc}. 
Many of these works rely extensively on a local reference frame that is assumed to be repeatably constructed on the point sets~\citep{petrelli2011repeatability,mian2010repeatability,Melzi_2019_CVPR}. 
With the advances in deep networks, these methods have been replaced by their learned counterparts~\citep{zeng20163dmatch,khoury2017CGF,deng2018ppfnet,deng2018ppffold, gojcic20193DSmoothNet,Choy2019FCGF}. Be it data driven or not, a large portion of all these works owe their robustness to the local reference frames unless the input is made invariant to rotations~\citep{deng2018ppffold,zhao2019quaternion}. One of the aims of this paper is to extend the findings regarding locally rigid LRFs to the non-rigid.

\paragraph{Case of Deformable Shapes}
For the non-rigid shapes, pointwise descriptors are mostly designed to be intrinsic in order to handle isometric deformations~\citep{aubry2011wave, rustamov2007laplace, sun2009concise} and scale~\citep{bronstein2010scale}. However, designing a descriptor by hand is a cumbersome task. It requires manual balancing of the trade-off between robustness and discriminability, and often relies on heuristics to capture local patterns. Learning based methods are well suited for this task, given the availability of enough training samples. Pioneering works in the field include \cite{bronstein2011shape} which extended a ``bag of features'' approach to non-rigid 3D shape retrieval, and \cite{litman2013learning} which utilized a Mahalanobis metric learning to optimize a parametric spectral descriptor for shape matching. The success of deep learning in computer vision, has motivated a new active research area termed \textit{Geometric Deep Learning}~\citep{bronstein2017geometric}. A main challenge in this field is how to construct basic operations such as convolution and pooling for geometric data structures. One line of work has opted for extracting geodesic patches~\citep{masci2015geodesic, boscaini2016learning, monti2017geometric}. This way, the convolution operator is defined intrinsically on the manifold guaranteeing invariance to isometric deformations. The challenge lies in constructing repeatable local patches with a canonical orientation. For nonrigid shapes these are often based on curvature, but other approaches exist such as ~\cite{huang2015repeatable} and the recently proposed GFrames~\citep{Melzi_2019_CVPR}. On the other hand, spectral techniques~\cite{bruna2013spectral, henaff2015deep} generalize a convolutional network through the Graph Fourier transform, thus avoiding the need for a patch. A polynomial parameterization of the learned filters was proposed in~\cite{defferrard2016convolutional} in order to spatially localize the kernel and reduce the learning complexity. Common to most learning based methods, is that they use precomputed descriptors as inputs and improve upon them through learned operations. Few works explored directly using raw mesh data. In \cite{poulenard2019multi} it was shown that using 3D coordinates underperform SHOT~\citep{tombari2010unique} as input; while in~\cite{verma2018feastnet} the opposite conclusion was reached. MeshCNN~\citep{hanocka2019meshcnn} defined convolution on triangular meshes treating the \textit{edge} as a first citizen, rather than the nodes. Using angles and edge ratios as input features they were able to work on the raw mesh while being rotation, translation and (uniform) scale invariant.






\subsection{Continuous convolution}
Key to our approach is a continuous convolution operator. Realizing that the input points are merely samples from an underlying continuous surface, this makes a much more natural formulation than treating the points as an unstructured cloud. Several works have explored the use of learned continuous kernels. In \cite{litany2016cloud, digne2014self} self similarities in the mesh patches were used via dictionary learning. More recently \cite{masci2015geodesic,boscaini2016learning} studied the extraction of local geodesic patches for constructing the equivalent of a convolutional neural network for 2-dimensional manifolds. A generalized form of these was introduced in \cite{monti2017geometric}. In \cite{atzmon2018point} a continuous processing of pointcloud was proposed by defining an extension operator that maps pointclouds to continuous volumetric functions. Another line of works parameterizes the continuous ambient function by another network. This concept was first introduced in \cite{jia2016dynamic}, where it was termed ``dynamic filter’’ as it allows to modify the convolution filters according to the input, instead of using fixed ones. In \cite{dai2017deformable} a deformation of the convolution kernel was used to dynamically react to the input image patch. A similar approach was taken for pointclouds in \cite{thomas2019kpconv}. Instead of dynamically modifying the kernel,~\cite{li2018pointcnn} proposed a $\chi-$transformation in order to to canonicalize the input points. Our continuous convolution resembles the most this line of work, however differently from pointcloud based dynamic filters we utilize the mesh structure to enrich our point features as described in Section \ref{par:cont_conv}.


\begin{figure*}
    \centering
    \includegraphics[width=\textwidth]{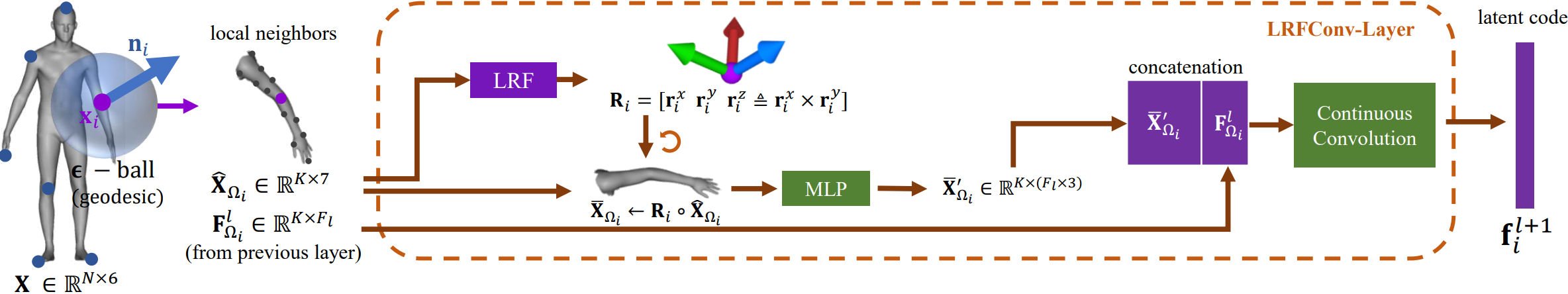}
    \caption{Our LRF-Conv layer. We treat a shape as a set of 3D coordinates and surface normals $\X\in\R^{N\times 6}$. Around each point $\x_i\in\R^6\equiv [x,y,z,n_x,n_y,n_z]^\top$ we consider a 3D geodesic local neighborhood $\Omega_i$: $\X^M_{\Omega_i}=\{\x_k \in \X \,:\, d(\x_i,\x_k)< \tau\}$, where $\tau$ is a threshold on the geodesic distance $d$. We then perform a farthest point sampling (FPS) on this set and retain $K$ neighbors. We denote this set $\X_{\Omega_i}=\text{FPS}(\X^M_{\Omega_i},K)$. We perform a de-mean operation, setting the coordinates of $\x_i$ as the local origin. We further augment each point with the geodesic distance to this new origin as well as the surface normal of the origin and let $\hat{\x}_{ij}\in\R^{7}=[\hat{x}_{ij},\hat{y}_{ij},\hat{z}_{ij},\hat{n}_{ij}^x,\hat{n}_{ij}^y,\hat{n}_{ij}^z,g_{ij}\triangleq d(\x_i, \x_{ij})]^\top$ represent the $j^\text{th}$ point in the local frame of $\x_i$. We use $\hat{\X}_{\Omega_i}=\{\hat{\x}_{ij}\}_{j\in\Omega_i}$ to refer to this augmented, centered local set and this is precisely the input to our LRF-Conv layer. Our layer takes as input a local reference frame~\citep{tombari2010unique} $\Rot_i$ assigned to the patch $i$. Next, we re-orient the input by $\Rot_i$ to get $\{\bar{\x}_{ij}\}_j\triangleq\bar{\X}_{\Omega_i}=\Rot_i\circ\hat{\X}_{\Omega_i}$. Note that the operator $\circ$ does not act on the distances $\{d_{ij}\}_j$ while rotating the rest. 
    We then use an MLP per entity (coordinates, normals, distances) to extend the input description to match the latent dimension and concatenate these with features extracted in the previous layer $\F_{\Omega_i}^l$, where $\F_{\Omega_i}^0 \triangleq \zero$. This concatenated feature is the input to our trainable continuous convolution yielding the latent features (output) of this $(l+1)^{\text{th}}$ layer. We show in green the learnable modules whereas purple depicts the data containers.
    }
    \label{fig:lrfconv}
\end{figure*}


\section{LRFConv Layers}
At the core of our contribution lies a continuous convolution layer that operates on a locally rectified point set and its geodesics. 
We call this the \emph{LRFConv} Layer and illustrate it in Figure \ref{fig:lrfconv}.
LRFConv receives a local patch $\hat{\X}_{\Omega_i}=\{\hat{\x}_{ij}\}_{j}=\{\hat{\x}_{i1}, \hat{\x}_{i2}\,,\,\dots\,,\,\hat{\x}_{iK}\}$ that is already centered on a given reference point $\x_i$ as input. 
This local patch is composed of a collection of those $K$ points lying in the region $\Omega_i$ being subsampled using a farthest point sampling algorithm~\citep{qi2017pointnet++,moenning2003fast}.
Along with its 3D coordinates, each $j^\text{th}$ point in this patch also carries two additional pieces of geometric information: the surface normal $\hat{\nb}_{{i}j}\in\Sp^3$ and the geodesic distance to the reference (center) point $d({\x}_i,{\x}_{ij})$, a quantity that is preserved under isometries. 
For a vertex $j$ centered around the $i^\text{th}$ vertex, this yields a 7-dimensional point representation: 
\begin{equation}
   \hat{\x}_{ij} \in \R^{7} = \begin{bmatrix} \hat{x}_{ij} & \hat{y}_{ij} & \hat{z}_{ij} & \hat{n}_{ij}^x & \hat{n}_{ij}^y & \hat{n}_{ij}^z & g_{ij}\triangleq d(\x_i, \x_{ij})
    \end{bmatrix}^\top
\end{equation}
where subscript ${ij}$ refers to the index of the $j^\text{th}$ point in the neighborhood of $i^\text{th}$ vertex: $j\in\Omega_i$. 
In order to build resilience among six degrees of freedom (DoF) rotations we re-orient this patch using a local reference frame. 
To this end, we first compute an LRF for all points in the vertex set $\X$. Then each patch $\hat{\X}_{\Omega_i}$ is assigned an LRF in accordance with its index.
The axes of this LRF are assembled into a rotation matrix $\Rot_i \in SO(3)$ which can be used to transform the patch to a canonical alignment: $\bar{\X}_{\Omega_i} = \Rot_i \circ \hat{\X}_{\Omega_i}$. 
Here the $\circ$ operator only acts on coordinates and normals separately.
We then consider the aligned local patch $\bar{\X}_{\Omega_i}\triangleq \{\bar{\x}_{ij} \in \R^7\}_j$. Note that after such transformations $\bar{\x}_i\equiv\mathbf{o}$, a constant vector.
We compute an intermediate feature representation for the whole patch using the entirety of the information collected up to this point and extend it with the help of three multi layer perceptrons (MLPs). 
We use one MLP per each of the coordinates, normals and geodesic distances in order to match the dimension of the latent features propagated from the previous layer, denoted as $\F_{\Omega_i}^l\in\R^{K\times F_l}$ where $F_l$ is the dimension of the features. This generates $\bar{\X}^\prime_{\Omega_i} \in \R^{K\times (F_l\times 3)}$.
Note that an essential quantity in feeding forward the information generated in the previous layers to the later layers of our network is $\F_{\Omega_i}^l$. 
Thus, we concatenate the output of the said MLPs with $\F_{\Omega_i}^l$ and feed the resulting matrix into a continuous convolution operation producing the output features of this layer $\fb_i^{l+1}\in\R^{F_{l+1}}$.
Note that in the beginning we initialize the features to zeros: $\F_{\Omega_i}^0 \triangleq \zero$.

In the following, we will first present the details of our LRF computation and then dig deeper into the continuous convolutions.

\paragraph{Local Reference Frame (LRF)}

In order to introduce invariance to translations and rotations as well as building robustness to noise, many of the handcrafted descriptors rely on the estimation of a local coordinate system that varies equivariantly with the global transformation of the object. 
We use a similar idea to endow our deep features with invariances. 
A frame of reference (LRF) can be parameterized as a rotation matrix $\Rot_i =[\rb^x_i,\rb^y_i,\rb^z_i] \in SO(3)$ where each column corresponds to an axis of the local coordinate frame. 
In our work, we switch between two LRFs depending on whether the data is real or synthetic. 
For scanned point sets, we use SHOT's LRF~\citep{tombari2010unique} thanks to its uniqueness and robustness to noise. 
The second kind that is suited to less noisy, synthetic meshes is inspired by Texturenet~\citep{huang2019texturenet}: The first axis is aligned with the surface normal at $\x_i$: $\nb_i$. 
The second axis is determined by the direction of maximum curvature projected on the tangent plane defined by the surface normal (first axis). 
The third and the final axis is simply the cross product of the two: $\rb^z_i=\rb^x_i\times \rb^y_i$. 
Such LRF construction reduces the degree of ambiguity from four as in~\citep{huang2019texturenet} to two. 

\paragraph{Continuous Graph Convolution}
\label{par:cont_conv}

\begin{figure*}[t]
    \centering
    \includegraphics[width=\textwidth]{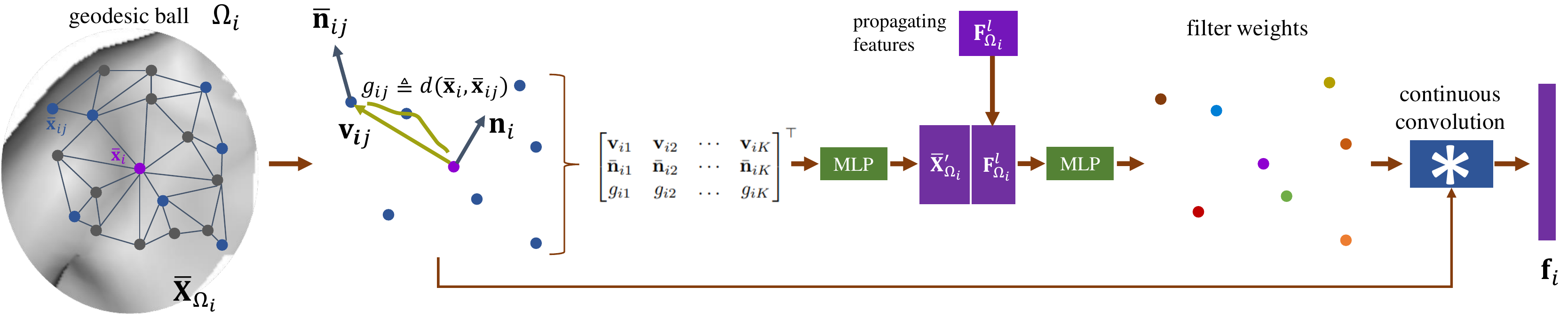}
    \caption{Details of our continuous geodesic convolution. The figure shows in more detail the computation of the input to the MLP and subsequently to the convolution in the unstructured domain. Our MLP that regresses the convolution kernel learns a mapping from the high dimensional point/patch representation to a matrix of weights.
    }
    \label{fig:lrfpipeline}
\end{figure*}

An important portion of the success of the CNNs is attributed to the 2D convolutions that are well suited to the structured grid underlying the pixels. 
Unfortunately, for unstructured 3D data, defining such a grid is not trivial and hence the primary tools for point set processing such as PointNet~\citep{qi2017pointnet} prefer to ignore the domain and apply point-wise convolutions. 
Yet, taking into account the neighborhood structure is shown to be advantageous~\citep{qi2017pointnet++}. 
Thanks to the availability of the mesh structure, we could define an unstructured convolution analogous to the 2D that considers the mesh surface. 
To this end, we use continuous graph convolutions., whose details we show in Figure \ref{fig:lrfpipeline} and explain below.

As discussed in the introduction of this section, we first extract a patch $\X^M_{\Omega_i}=\{\x_k \in \X \,:\, d(\x_i,\x_k)< \tau\}$ according to the geodesic distance $\tau$ of the reference point $\x_i$.
The we use FPS to get K points within $\X^M_{\Omega_i}$ and center them given $\x_i$ to get $\hat{\X}_{\Omega_i}$.
After that, we rotate $\hat{\X}_{\Omega_i}$ using the LRF to obtain $\bar{\X}_{\Omega_i}$. $\bar{\X}_{\Omega_i}$ carries four feature components per each point in the neighborhood of $\x_i$: the coordinate $\vb_{ij}$, the aligned normal $\bar{\nb}_{ij}$, the geodesic distance $g_{ij}$, and the feature from previous layer $\fb_{ij}^{l}$.
They could be expressed as $\vb_{ij}=\bar{x}_{ij}-\bar{x}_{i}$, $\bar{\nb}_{ij} = \Rot_i \hat{\nb}_{ij}$, and $  g_{ij}=d(\bar{\x}_i, \bar{\x}_{ij})$.
Note that $\fb_{ij}^{l}$ is a rotation invariant feature.
We will justify choosing these four feature components in Section \ref{sec:ablation_study}.
In our implementation, the first layer omits this feature. Though, signals such as color could be whenever available.
For the following layers, $\fb_{ij}^{l} (l > 0)$ is the computed rotation invariant feature.
If $\fb_{ij}^{l}$ exists, then we first use three MLPs ($MLP_{vb}$, $MLP_{nb}$, and $MLP_{g}$) to expand the dimension of $\vb_{ij}$, $\bar{\nb}_{ij}$, and $g_{ij}$ to match the dimension of $\fb_{ij}^{l}$.
If $\fb_{ij}^{l}$ does not exist, we expand the dimension of $\vb_{ij}$ to 9.
This can help making the network treat each feature's components equally in the next step.
After concatenating the expanded input descriptions $\bar{\x}^\prime_{ij}$ and $\fb_{ij}^{l}$, we use another $MLP_{w}$ to regress a weight matrix whose size is is $F_l \times F_{l+1}$. Subsequently, we apply a discrete convolution operation between this convolution weights matrix and the input features:
\begin{equation}
    \fb_i^{l+1}=\sum_{j=1}^{K}MLP_{w}([\bar{\x}^\prime_{ij}, \fb_{ij}^{l}])[\bar{\x}^\prime_{ij}, \fb_{ij}^{l}]
    \label{eq:conv}
\end{equation}
With that we update the feature at the reference points $\x_i$. The whole process is depicted in Figure \ref{fig:lrfpipeline}.


\section{Network Architecture}
\begin{figure*}[t]
    \centering
    \includegraphics[width=\textwidth]{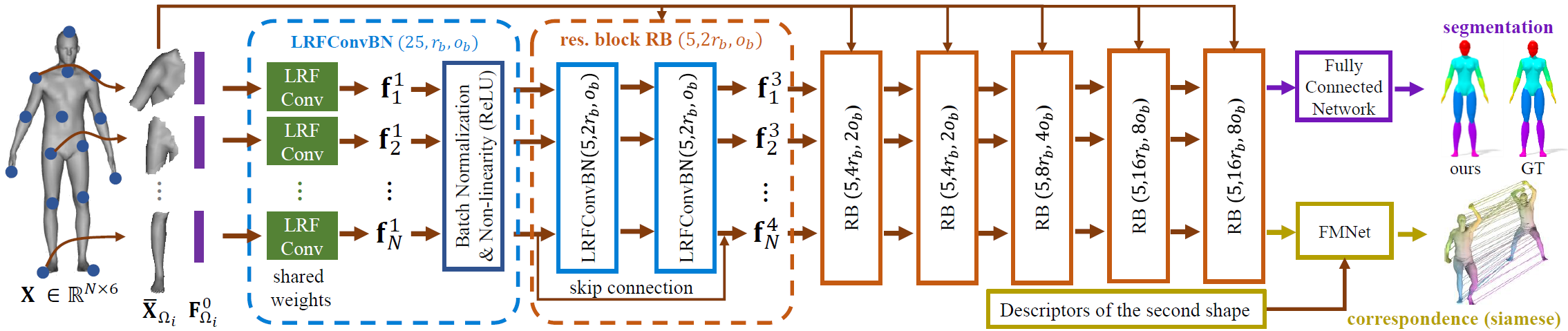}
    \caption{Entirety of our architecture. We extract a local patch centered around each point by querying a geodesic neighborhood. These patches are sent into a sequence of LRFConv layers followed by a batch normalization (BN) and ReLU non-linearity. We also add skip connections to be able to increase the depth and avoid the vanishing gradients. After 13 layers of LRFConv we arrive at our latent features which can be used to address common tasks such as human body segmentation or correspondence estimation. When LRFConv is followed by a BN and ReLU, we call this an \textit{LRFConvBN} layer and parametrize it by three respective arguments: $K$ the number of points in the patch, $r_b=0.003$ the base radius that determines the size of the neighborhood, $o_b=32$ where $o_l=\lambda o_b$ sets the dimension of the output features. The residual blocks (RB) that are composed by two LRFConBNs are similarly parametrized. Note that for correspondence estimation we have a weight sharing siamese architecture where the latent features of a paired shape are fed into a deep functional map network~\citep{litany2017deep} along with the features of the base (current) mesh.
    }
    \label{fig:network_arch}
\end{figure*}

Our network architecture as shown in Figure \ref{fig:network_arch} consumes each local part separately and involves stacking of the LRF-Conv layers with skip connections. 
The upper skip links marked in brown denote that the coordinates, normals, and geodesic distances are also fed forward.
In addition, by speaking of LRFConvBN in Figure \ref{fig:network_arch}, it means a LRFConv layer followed by a batch normalization layer \cite{DBLP:journals/corr/IoffeS15} ,and a non-linear layer, in our case ReLU \cite{nair2010rectified}.
The part of network before branching into task specific modules is what we call the \textit{learned shape descriptor (LSD)}.
The architecture of LSD is motivated by ResNet~\citep{he2016deep}.
As the layers deepen, we gradually increase the perception field of the network and the length of the feature while having skip connections to increase the depth of the network. This avoids the vanishing gradient problem.
Once the LSD is extracted for each vertex (local patch) on the shape, we use it for two tasks, human body segmentation and shape correspondence. 


\paragraph{Part Segmentation}
\label{paragraph: network architecture part segmentation}

The fully connected network for part segmentation is composed of 7 residual blocks. 
For each block the dimension of the output feature is reduced by half with the first dimension being 512.
In the last residual block, the dimension equals to the number of classes $M$ which is 8. 
And a softmax layer is followed.
We get an output label $\svec_i$ for each patch anchored at $\x_i$.
For the segmentation task, we minimize the cross-entropy loss between the output predictions $\{s_i\}$ and the ground truth segmentation labels $\{y_i\}$.

\begin{equation}
    \ell = -\frac{1}{M} \sum_{i=1}^{M} y_i \log s_i
\end{equation}

\paragraph{Correspondence Estimation}
The fully connected network for estimating correspondence consists of 7 residual blocks.
In order to have a fair comparison with FMNet~\citep{litany2017deep}, the dimension of the output / feature of each block is set to 352. This is identical to the length of the latent feature used in FMNet.
We use the shared weights of LSD followed by the fully connected network to extract the feature from the target shape $\mathcal{Y}$ and the source shape $\mathcal{X}$. Then we follow the loss function proposed in FMNet:
\begin{equation}
    \ell=\frac{1}{|\mathcal{X}|}||\Pb\circ (\mathbf{D}_{\mathcal{Y}} \Pi^{*})||_{F}^{2}
\end{equation}
Note that our correspondence estimation approach resembles FMNet's. However, in addition to our continuous convolutions, we avoid using the handcrafted SHOT descriptors and replace them with LSDs.

\section{Experimental Evaluation}

\subsection{Datasets}
\label{sec:datasets}
We demonstrate the efficacy of our learned descriptor on two cornerstone tasks in shape analysis: dense shape correspondence and part segmentation. To this end, we utilize two datasets. 
\paragraph{Part Segmentation.}
For the segmentation task, we use the human segmentation benchmark introduced in \cite{ToricCNN2017}. This dataset consists of 370 models fused from multiple human shape collections: SCAPE \citep{anguelov2005scape}, FAUST \citep{bogo2014faust}, MIT animation datasets \citep{vlasic2008articulated} and Adobe Fuse \citep{adobe_fuse}.
All models were manually segmented into eight parts as prescribed by \cite{kalogerakis2010learning}. 
The test set is the 18 human models from the SHREC dataset \citep{giorgi2007shape}.
The variety of data sources makes the problem especially challenging as each collection has a different sampling and appearance. Moreover, the SHREC dataset was used solely for testing which calls for a high generalization ability. 

\paragraph{Shape Correspondence.} To showcase our descriptor learning module in the task of shape matching we use the FAUST dataset \citep{bogo2014faust}. The data contains 100 human high resolution scans belonging to 10 different individuals at 10 different poses each. The scans were all registered to a parametric model with 6890 vertices and consistent triangulation. We call this set ``Synthetic FAUST''. We also test our method on the more challenging set of the original scans. 
\subsection{Part Segmentation}
\label{sec:part_seg}
Given an input mesh we use our network to predict for each vertex the part segment it belongs to. 
At train time, we select 2000 random points from each mesh as input to the network, and use the corresponding segmentation label as the supervision signal.
We train our network for 200 epochs. In all our experiments we use the Adam optimizer~\citep{kingma2014adam} with a fixed learning rate of $10^{-3}$, $\beta_{1}=0.9$, $\beta_{2}=0.999$, and $\epsilon=10^{-8}$.
We compare our results with the two variations of MDGCNN~\citep{poulenard2019multi} as proposed by the authors, using either raw 3D coordinates or precomputed SHOT~\citep{tombari2010unique} descriptors. 
To better understand the influence of the continuous convolution module (CC) we also compare with a simplified version of our pipeline, where the continuous convolution is replaced by a standard PointNet (PN)~\citep{qi2017pointnet}. 
Our results are summarized in Table~\ref{tab:segmentation}. 
It can be seen that our method in both of its forms outperforms MDGCNN, while using the continuous convolution further boosts the performance. 

\begin{SCtable}
\centering
\tablestyle{3pt}{1.2}
\begin{tabular}[b]{l|c|c}
    Method & Input feature & Accuracy \\
    \shline
    MDGCNN~\citep{poulenard2019multi} & 3D coords & 88.61    \\
    MDGCNN~\citep{poulenard2019multi} & $\text{SHOT}_{12}$ & 89.47    \\
    \hshline
    Ours PN & 3D coords  & 89.69    \\
    Ours CC & 3D coords  & {\bf 89.88}    \\
\end{tabular}
\caption{
    We compare our results with the two variations of MDGCNN~\citep{poulenard2019multi} as proposed by the authors, using either raw 3D coordinates or precomputed SHOT~\citep{tombari2010unique} descriptors. 
    To better understand the influence of the continuous convolution module (CC) we also compare with a simplified version of our pipeline, where the continuous convolution is replaced by a standard PointNet (PN)~\citep{qi2017pointnet}.
}
\label{tab:segmentation}
\end{SCtable}

\begin{figure*}
    \centering
    \includegraphics[width=\textwidth]{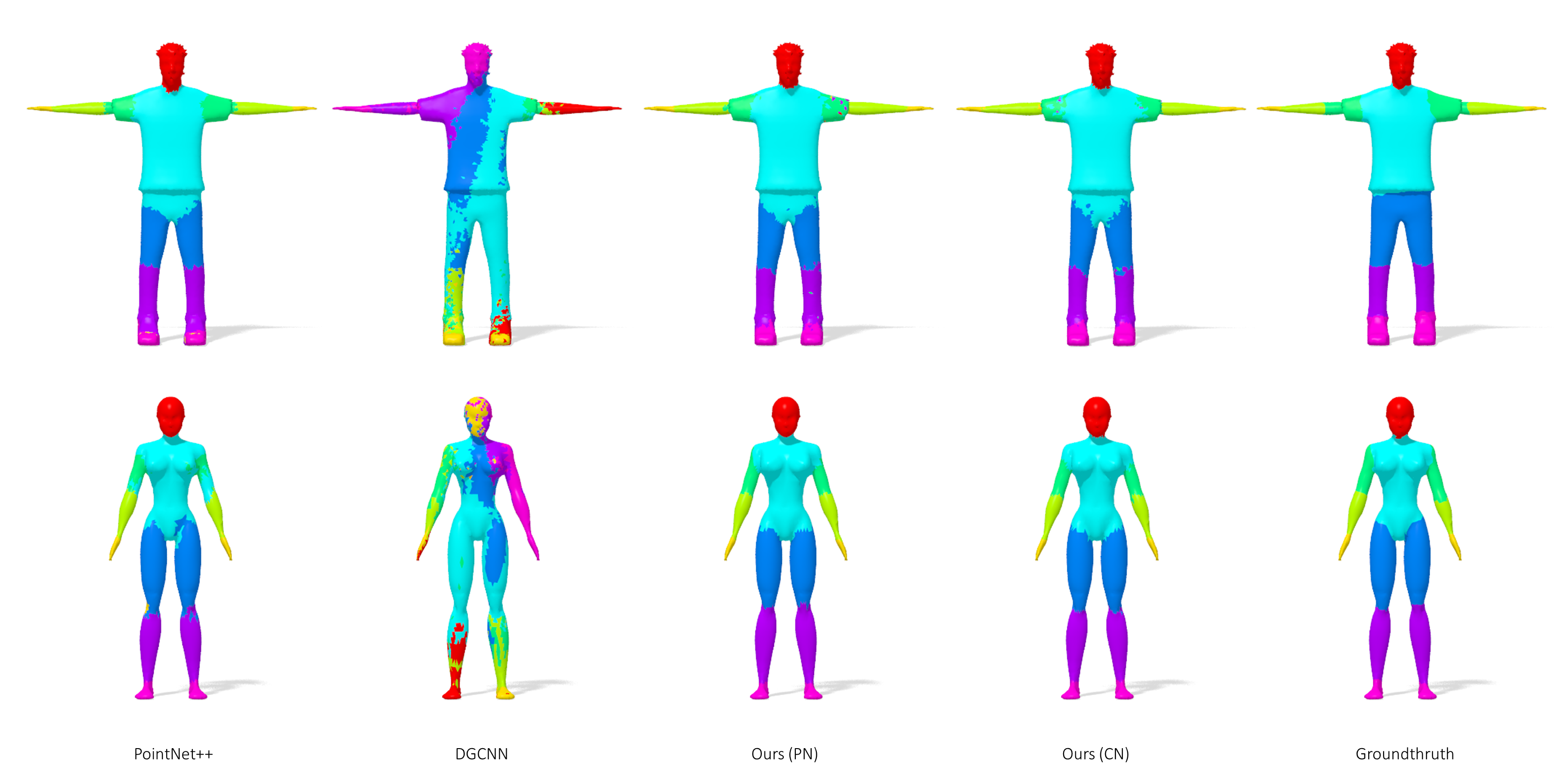}
    \caption{
        A qualitative evaluation of our method on human body segmentation comparing to other learning-based approaches.
    }
    \label{fig:segmentation}
\end{figure*}

Importantly, we achieve this by using raw 3D coordinates as input and by which bridges the gap reported in MDGCNN allowing to remove the dependence on manually designed features. This desired behaviour is expected since our network imitates the design philosophy of SHOT.
We further present a qualitative evaluation of the part segmentation in Figure~\ref{fig:segmentation}.

\subsection{Rotation invariance}
As described in the introduction, depending on the dataset and task at hand, a descriptor should be invariance to different transformations. In the case of human body models and part segmentation it is natural to ask for invariance to rigid transformations and articulations. While the latter is achieved through learning from examples, the former is taken care of by construction using our proposed LRF. Baking invariance into the descriptor by construction is not only natural to the problem setting but also more sample efficient. To see this, we take two baseline methods which are not rotation invariant, and train them with and without  rotation augmentation. We also create an augmented test set, where each shape has been rotated in 128 different angles. We summarize the results of these methods and ours in Table~\ref{tab:rotation_inv}. 
As can be seen, when the non rotation invariant networks are tested on rotated examples their performance drops significantly. Adding rotation at train time helps shrinking the performance gap however, both still under perform compared to our rotation invariant solution.

\begin{figure*}[b!]
    \centering
    \includegraphics[width=\textwidth]{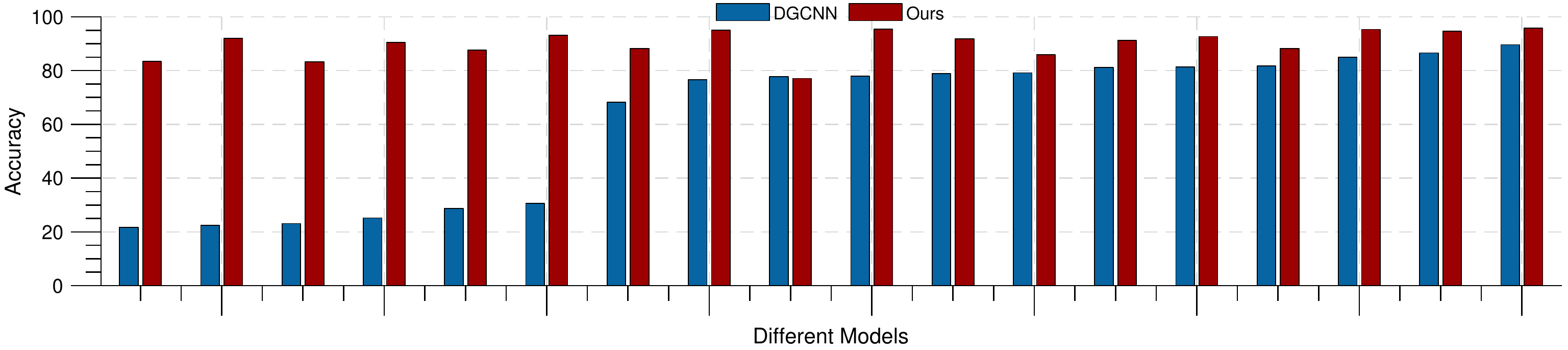}
    \caption{
    We compare the results of our network with DGCNN on each of the 18 test shapes.
    To better visualize the effect of the rotation angle we sort the test samples (x-axis) according to the performance of DGCNN. 
    This clearly shows a gap between samples where the input shape was lying down (first 6 examples) and the ones which were upright. }
    \label{fig:per_example_seg_acc}
\end{figure*}

Looking closely at the results of PointNet++~\citep{qi2017pointnet++} one notices that the results improve on the non-rotated test set by adding the augmentation at train time. This is explained by the fact that the models in both train and test sets are either upright or lying down but with very different distribution between the two poses. What is less intuitive is why this improvement is not achieved for DGCNN. From our experiments we conclude that the method could not benefit from the augmentation and instead reduced to solving the average case, perhaps due to limited capacity. 
In Figure~\ref{fig:per_example_seg_acc} we compare the results of our network with DGCNN on each of the 18 test shapes. 
To better visualize the effect of the rotation angle we sort the test samples (x-axis) according to the performance of DGCNN. 
This clearly shows a gap between samples where the input shape was lying down (first 6 examples) and the ones which were upright. 


\begin{table*}[t!]
\centering
\tablestyle{3pt}{1.2}
\begin{tabular}{c|c|c|c|c|c}
Train with rotation & Test with rotation & PointNet++ & DGCNN & Ours(PN) & Ours(CC)       \\
\shline
Yes      & Yes  &   85.35    & 43.88 & \multirow{4}{*}{89.69}   & \multirow{4}{*}{\textbf{89.88}}            \\
Yes      & No   &   85.85    & 36.99 &                          &               \\
No       & Yes  &   56.95    & 36.65 &                          &               \\
No       & No   &   75.81    & 66.35 &                          &
\end{tabular}
\vspace{2mm}
\caption{The accuracy on human body segmentation of different learning-based approaches under training and test with and without rotation augmentation}
\label{tab:rotation_inv}
\vspace{-1mm}
\end{table*}
\subsection{Shape matching}
Finding dense shape correspondences between a pair of shapes is one of the most important and most explored problems in shape analysis. As described earlier, state of the art methods utilize hand crafted descriptors as input. Here, instead, we propose to learn the descriptor directly from raw 3D coordinates by utilizing our proposed LRFConv. Our descriptor can in principle be combined with any matching pipeline. In this work we make use of FMNet~\citep{litany2017deep}, one of the best performing methods in the task of shape matching. Specifically, it accepts a pair of shapes as input together with their computed Laplacian eigenfunctions and per-point features. Then, both shape features are passed through a (siamese) feed forward network to get refined descriptors. These, in turn, are used to compute a functional map aligning the shape eignefunctions. Finally, these are used to predict a point-to-point soft-correspondences which are converted to matchies by taking the maximal probability per point. In the original work of ~\citep{litany2017deep} and its unsupervised follow up~\citep{halimi2018self} SHOT descriptors were used. Here, instead, we replace it with our learned descriptors and train the network in an end-to-end fashion. 

\begin{figure}[bt]
\begin{minipage}{1.0\linewidth}
    \centering
    \input{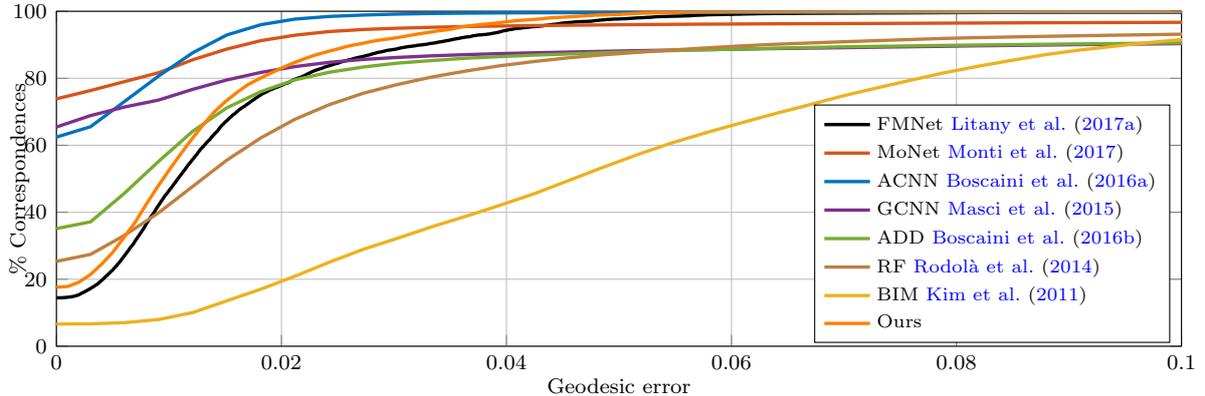}
\end{minipage}
\caption{Comparison with learning-based approaches on the FAUST humans dataset. By using our proposed descriptor, we were able to improve upon the results of FMNet with SHOT.
}
\label{fig:kimcurve}
\end{figure}

\paragraph{Synthetic FAUST}
We first demonstrate our performance on the synthetic FAUST dataset described in \ref{sec:datasets}. We follow the evaluation protocol as prescribed by ~\cite{monti2017geometric} were the 100 models are split into 80 train and 20 test shapes, and the matching is performed with respect to a single fixed null shape. 
We train the network for 200 epochs using the same optimization hyper parameters as described in \ref{sec:part_seg}. The results are summarized in Figure ~\ref{fig:kimcurve}. 
As can be seen, by using our proposed descriptor we were able to improve upon the results of FMNet with SHOT. 
We include the performance of other methods for the sake of completeness. 

\begin{figure*}
    \centering
    \includegraphics[width=\textwidth]{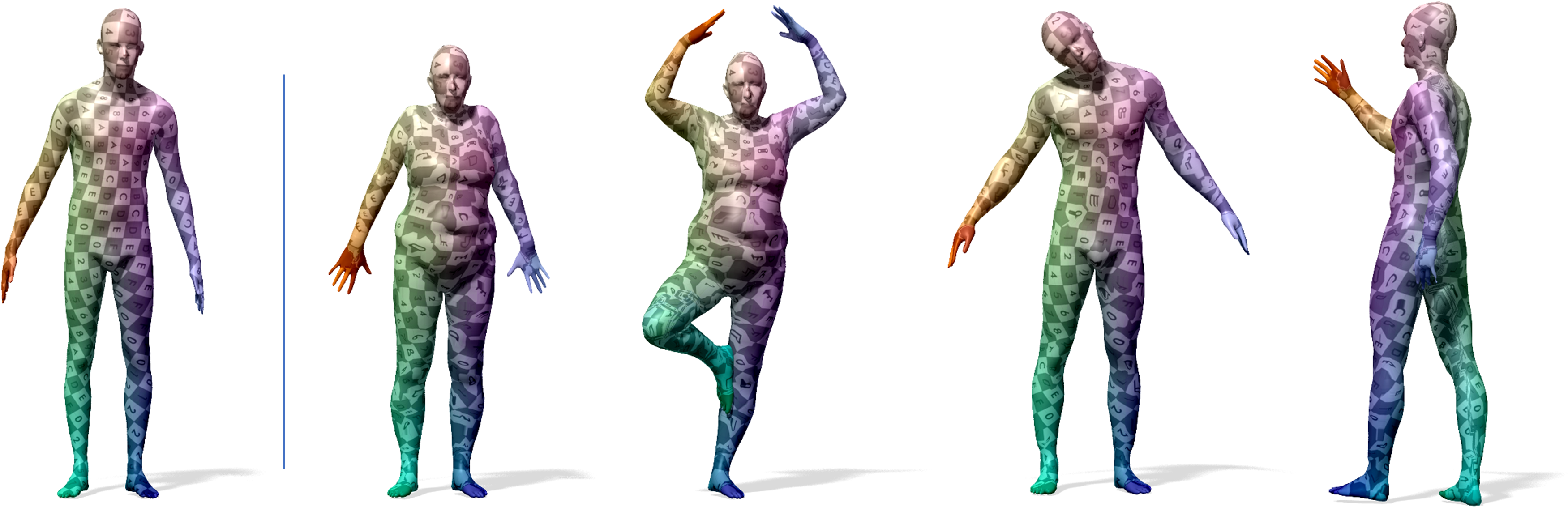}
    \caption{A qualitative evaluation of our matching results on synthetic FAUST. On the left of the vertical border is a textured target shape. To the right are four source shapes from the test set with their texture pulled back from the target according to the recovered correspondences.}
    \label{fig:matching_accuracy}
\end{figure*}

In Figure ~\ref{fig:matching_accuracy} we show a qualitative evaluation of our matching results. 

%




\subsection{Ablation Study}
\label{sec:ablation_study}

\begin{wraptable}{r}{6.0cm}
    \tablestyle{3pt}{1.2}
    \begin{tabular}{l|c}
        Network                     & Accuracy \\
        \shline
        Full                        & 98.30    \\
        Replace CC with PN          & 98.29    \\
        Remove $g_{ij}$             & 98.24    \\
        Remove $\nb_{ij}$           & 97.98    \\
        Remove LRF                  & 95.24    \\
        Remove $\fb_{ij}^{l}$       & 94.40    \\
        Remove $\vb_{ij}$           & 75.21    
    \end{tabular}
    \vspace{2mm}
    \caption{Ablation study on the design choices of our network ingredients. By removing different components and retraining we evaluate their importance.
    }
    \label{ablation_study}
    \vspace{-1mm}
\end{wraptable}


In this section we provide analysis of the design choices made when constructing our descriptor learning network. To this end, we utilize the synthetic FAUST dataset and the task of part segmentation. 
%
%
Since we are working in a simplified setting with only one dataset, we make two modifications to the network. First, we reduce the base feature dimensionality (see $o_b$ in Figure \ref{paragraph: network architecture part segmentation}) from 32 to 4. Second, since the number of vertices is kept fixed we use all vertices instead of subsampling a 2K subset. 
To evaluate the importance of each ingredient, we retrain and evaluate the network performance with and without it. 
Results are summarized in Table ~\ref{ablation_study}.
It can be seen that the accuracy achieved with both \ourscg{} and \ourspg{}, is very high. 


The importance of the rest of the components in descending order is: the coordinates $\vb_{ij}$, removing feature propagation from previous layers $\fb_{ij}^{l}$, LRF, normals $\nb_{ij}$, and geodesic distance to the central vertex $g_{ij}$.
As expected, the coordinate $\vb_{ij}$ is playing the biggest role as it is holds the most information of the local patch geometry. 
Another motivation to include normals comes from the results reported in \cite{koch2019abc}, where it was shown that all state-of-the-art networks struggle with accurately estimating the normal of a patch. The results justify the inclusion of each of the components.

\section{Conclusion}
In this work we have studied the problem of learning shape descriptors. A main motivation for this work was the fact that sate-of-the-art learning based techniques were still relying on hand crafted descriptors. Here, we showed that this is mainly due to the usage of the LRF. By baking the computation of an LRF into the design of the network we were able to bridge the gap and outperform manual descriptor- based methods with using raw mesh features: coordinates, normals, and geodesic distances. In addition, we introduced a continuous convolution kernel which allows the filters to dynamically react to the input features. We demonstrated the performance of our proposed method on two important tasks: shape matching and part segmentation. Albeit the usage of a continuous convolution, current method including ours, still rely heavily on the set of sampling points and the sampling method (FPS in our case). This is of course an unwanted behaviour as the result should depend on the underlying surface. In future work we plan to explore this direction. 

\bibliography{references}

\end{document}